\documentclass{article}
\usepackage{spconf,amsmath,graphicx}
\usepackage{amssymb}
\usepackage{booktabs}
\usepackage{multirow}
\usepackage{listings}
\usepackage{adjustbox}
\usepackage{url}
\usepackage{makecell}
\usepackage{cite}
\usepackage{bigdelim}

\usepackage[T1]{fontenc}

\usepackage[labelformat=simple]{subcaption}

\usepackage[font=small,labelfont=bf]{caption}

\usepackage[dvipsnames]{xcolor}

\renewenvironment{thebibliography}[1]{
  \begin{oldthebibliography}{#1}
    \setlength{\itemsep}{0em}
    \setlength{\parskip}{0em}
}
{
  \end{oldthebibliography}
}
\usepackage{setspace}

\allowdisplaybreaks

\title{Updated Corpora and Benchmarks for Long-form Speech Recognition}
\name{Jennifer Drexler Fox$^1$, Desh Raj$^2$, Natalie Delworth$^1$, Quinn McNamara$^1$, Corey Miller$^1$, Migüel Jetté$^1$}
\address{$^1$Rev.com; $^2$Center for Language and Speech Processing, Johns Hopkins University, USA.}
\begin{document}
%
\maketitle
\begin{abstract}
The vast majority of ASR research uses corpora in which both the training and test data have been pre-segmented into utterances. In most real-word ASR use-cases, however, test audio is not segmented, leading to a mismatch between inference-time conditions and models trained on segmented utterances. In this paper, we re-release three standard ASR corpora\textemdash TED-LIUM 3, Gigapeech, and VoxPopuli-en\textemdash with updated transcription and alignments to enable their use for long-form ASR research. We use these reconstituted corpora to study the train-test mismatch problem for transducers and attention-based encoder-decoders (AEDs), confirming that AEDs are more susceptible to this issue. Finally, we benchmark a simple long-form training for these models, showing its efficacy for model robustness under this domain shift.
\end{abstract}

\vspace{-2em}
\section{Introduction}
\label{sec:intro}
\vspace{-0.5em}

%
Most ASR research uses corpora in which both the training and test data have been pre-segmented into utterances. 
Real-world audio, on the other hand, occurs as \textit{long-form} unsegmented recordings, leading to a mismatch between inference-time conditions and models trained on segmented utterances. 
This mismatch problem for long-form ASR has been well established in the literature~\cite{narayanan2019recognizing, lu2021input}, and researchers have sought to tackle it through better segmentation~\cite{huang2022e2e,Shu2023ACS}, large context acoustic modeling~\cite{Gong2022LongFNTLS,Hori2020TransformerBasedLE}, or rescoring with appropriate language models~\cite{Xiong2018SessionlevelLM,Chen2023LargescaleLM}. 
%

\begin{figure}[t]
    \centering
    \includegraphics[width=0.95\linewidth,trim={0cm 0.5cm 0cm 0.5cm},clip]{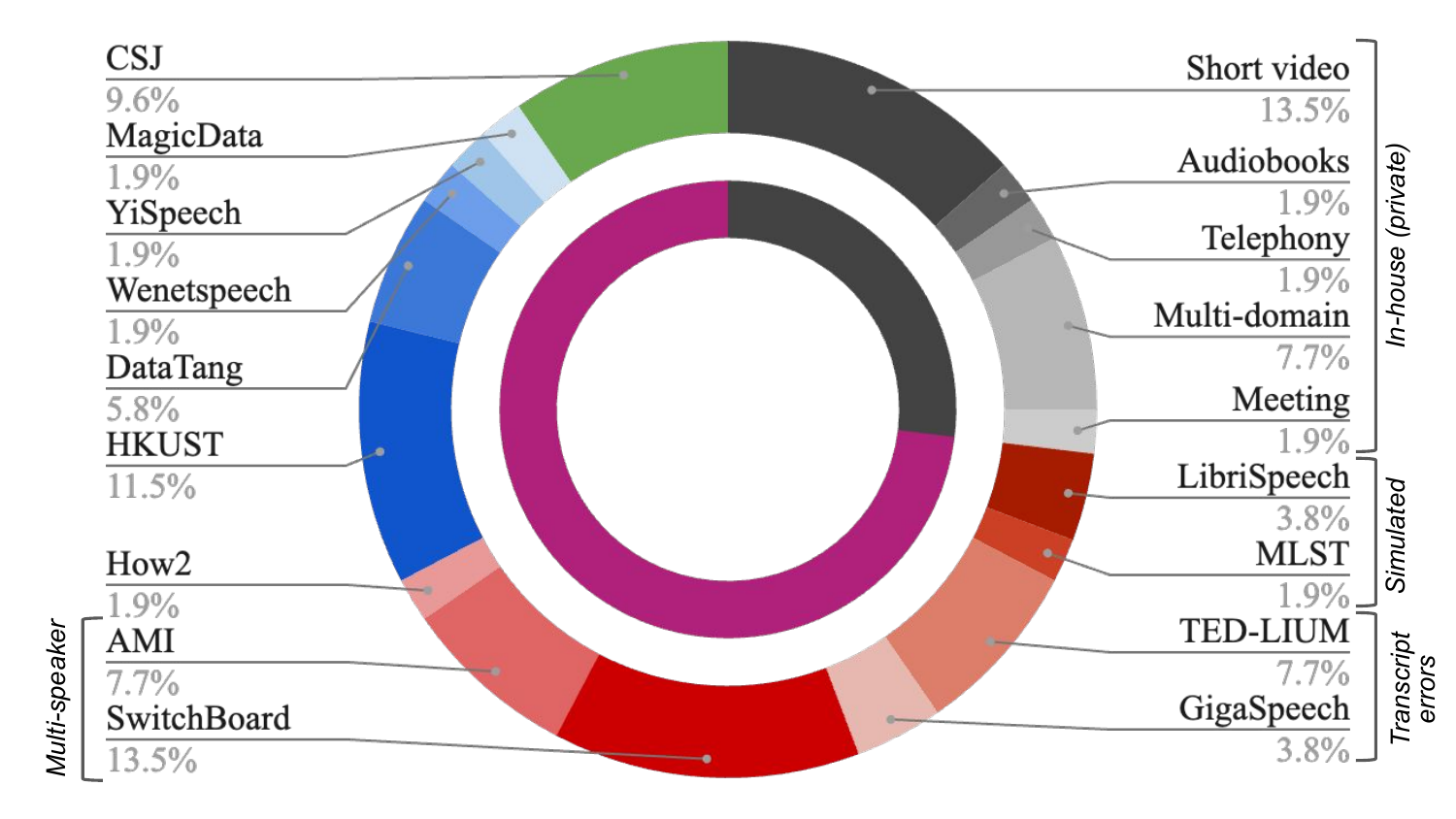}
    \caption{Statistics of in-house (gray) and public (colored) datasets used in long-form ASR research. Color shades represent languages: \textcolor{BrickRed}{English}, \textcolor{Blue}{Mandarin}, and \textcolor{OliveGreen}{Japanese}.}
    \label{fig:lf_stats}
\end{figure}

A significant fraction of these long-form modeling techniques have only been evaluated on in-house or simulated data.
In Fig.~\ref{fig:lf_stats}, we present corpus statistics from 36 published papers\footnote{These papers were manually selected based on an approximate depth-first search on the citation graph of a few seed papers, such as \cite{chiu2019comparison}.} on long-form ASR, showing that 26.9\% and 5.7\% used in-house and simulated data, respectively.
Even when publicly available ``true'' long-form corpora were used, they were often multi-speaker (21.2\%; e.g. AMI~\cite{Carletta2005TheAM} and SwitchBoard~\cite{Godfrey1992SWITCHBOARDTS}), non-English (32.6\%), or contained missing segments (11.5\%; GigaSpeech and TED-LIUM).
This obscures the real long-context modeling problem with orthogonal issues such as overlapped speech, tokenization, or incorrect evaluation.
%

To enable fundamental research on this problem, we release long-form versions of three English ASR corpora: TED-LIUM 3~\cite{Hernandez2018TEDLIUM3T}, Gigapeech~\cite{Chen2021GigaSpeechAE}, and VoxPopuli-en~\cite{Wang2021VoxPopuliAL}. 
Although the original releases for these datasets provide full recordings, the completeness of their transcriptions varies significantly, thus creating several challenges towards their use for long-form ASR.
For instance, several portions of the recording may be untranscribed, or some segments may have been removed due to alignment problems or non-verbatim transcription.
%
We reconstitute these long-form corpora through \textit{linking} and \textit{expansion} techniques (Section~\ref{sec:method}).

Finally, we use these reconstituted corpora to demonstrate the train/inference mismatch problem using baseline ASR models trained on the original short-form segments, for both transducers~\cite{Graves2012SequenceTW} and attention-based encoder-decoders (AEDs)~\cite{Chorowski2015AttentionBasedMF}. 
We show that incorporating long-form training can significantly improve performance when using chunk-wise overlapped inference.
Our reconstituted versions of the corpora, along with word-level alignments, are publicly available as Lhotse manifests~\cite{zelasko2021lhotse}.\footnote{\url{https://github.com/revdotcom/speech-datasets}}

\vspace{-1em}
\section{Related Work}
\label{sec:related}
\vspace{-0.5em}

\begin{figure*}[t]
\centering
\includegraphics[width=\linewidth]{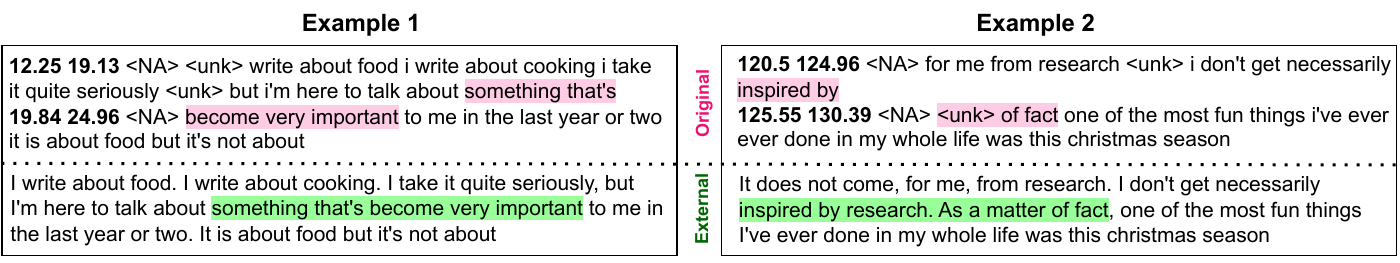}
\vspace{-1em}
\caption{Linkability determined by reference to external transcription}
\label{fig:example}
\vspace{-1.5em}
\end{figure*}

There is a large body of work addressing long-form ASR from several perspectives.
On the modeling front, researchers have extended conventional end-to-end ASR for large context handling through strategies such as: using history (or contextual) utterances to modify the encoder representation~\cite{Gong2022LongFNTLS,Masumura2019LargeCE,Kim2018DialogContextAE}), context expansion through preceding audio~\cite{narayanan2019recognizing,Hori2020TransformerBasedLE,Schwarz2020ImprovingRA}, and summarizing context through embeddings~\cite{Cui2023TowardsEA,Wei2022LeveragingAC}.
Chiu et al.~\cite{chiu2019comparison} compared popular end-to-end models for long-form ASR, finding that transducers are more robust than AEDs to the train-test mismatch.
%
%
Often, this mismatch can be partially alleviated through techniques such as random utterance concatenation~\cite{Lin2022RandomUC}, minimum word error rate (MWER) training~\cite{lu2021input}, and strong regularization~\cite{Chiu2020RNNTMF}.
OpenAI's Whisper model \cite{radford2023robust} takes a simpler approach to match training and inference conditions: in both cases, all audio is segmented into 30s chunks without any external VAD or diarizer. 

Chunk-wise overlapped inference~\cite{chiu2019comparison} is commonly used for offline decoding of long recordings. 
%
%
The related problem of segmentation has been addressed by using CTC-predicted blanks~\cite{Yoshimura2020EndtoEndAS}, a jointly trained continuous-integrate-and-fire (CIF) module~\cite{Shu2023ACS}, or using special tokens to predict segment boundaries~\cite{huang2022e2e}.
%
%
Language models (LMs) trained with expanded context have been used in first-pass decoding~\cite{Chen2020LSTMLMWL}, or more commonly for second-pass rescoring~\cite{Xiong2018SessionlevelLM,Chetupalli2020ContextDR,Irie2019TrainingLM,Chiu2021CrosssentenceNL,Chen2023LargescaleLM}.
%

Despite such interest, there is little consensus about best practices for training/decoding in long-form ASR, partially because of a lack of common benchmarks.
Although Earnings21~\cite{Rio2021Earnings21AP} and Earnings22~\cite{Rio2022Earnings22AP} were proposed to bridge this gap, they do not have any training data included, which makes it difficult to perform controlled investigations.

\vspace{-1em}
\section{Reconstituting long-form data}
\label{sec:method}
\vspace{-0.5em}

Our premise is that a ``true'' long-form corpus has long audio files and accompanying transcriptions. 
We used GigaSpeech, TED-LIUM, and VoxPopuli (\texttt{en} subset) as the base datasets for long-form reconstitution, since they provide such long recordings. 
These corpora have \texttt{train}, \texttt{dev}, and \texttt{test} partitions, but they are based on short segments and transcriptions (cut from the original recordings).
%
%
%
In this section, we describe our reconstitution process for converting an eligible long-form corpus into a true long-form corpus. 
This process has two possible realizations, \textit{linking} and \textit{expansion}. 
We view linking as a long-form repackaging of an existing corpus, whose results are directly comparable with results on the original corpora. 
In contrast, we view expansion as a new version of an existing corpus, since we have added new audio segments or transcriptions to the existing data. 
%
%

\vspace{-1em}
\subsection{Linking}
\label{sec:linking}
\vspace{-0.5em}

We define linking as concatenating original segments to make longer ones if  no speech or transcriptions lie in between. 
%

%
GigaSpeech comes with sequentially numbered segments that can be joined with previous and following segments when available.
In some cases, segments were missing in the sequence, and thus were presumed to be untranscribed and would not be able to be linked across. 

In contrast, internal resources were insufficient to allow for linking TED-LIUM. 
We observed several words in the audio that were not included in the transcriptions.\footnote{Previous work on long-form ASR using TED-LIUM seems to have missed or ignored this issue~\cite{Yoshimura2020EndtoEndAS}.} 
Most of these missing transcriptions were, however, present in the transcripts from a scrape of \url{ted.org}.\footnote{\url{https://www.kaggle.com/datasets/thegupta/ted-talk}}
Mapping TED-LIUM talks to the scrape was largely automatic, but a remainder of files needed to be associated by a semi-automatic method. 
By referring to these externally-sourced complete transcriptions, we were able to link adjacent segments in the original partitions when there was no missing text between the segments.
%
%
Fig.~\ref{fig:example} shows two representative segment pairs. 
In Example~1, the external transcriptions (at bottom) indicate that there were no missing transcriptions in between, so linking is possible. 
In Example~2, the external transcriptions indicate that there were in fact missing transcriptions between the segments and thus they cannot be linked, unless the corpus is expanded. 
We will describe this expansion process in the following section.

\vspace{-1em}
\subsection{Expansion}
\vspace{-0.5em}

Expansion is an optional process involving the addition of speech and/or transcriptions to an existing corpus. 
%

%
In VoxPopuli, purportedly exhaustive transcriptions are present in the original release, but 57\% of transcribed segments are not in the partitions. Segments were marked invalid when an ASR system got >20\% CER. 
We listened to several of these ``invalid'' segments and decided that their audio quality was not markedly different from other segments.
%
For any paragraph used in a particular partition, we resuscitated formerly invalid segments, allowing longer sequences to be reconstituted.

%
For TED-LIUM, expansion involved using the scraped transcriptions described in \ref{sec:linking} to replace the original transcriptions which had gaps.
%

%

\vspace{-1em}
\subsection{Statistics of reconstituted data}
\vspace{-0.5em}

\begin{table}[t]
\centering
\vspace{-0.5em}
\caption{Statistics of reconstituted data: total size of the set (HH:MM) and average segment length (seconds). $^\dagger$For long-form TED-LIUM, the numbers outside and within parentheses represent linked and expanded versions, respectively.}
\label{tab:stats}
\adjustbox{max width=\linewidth}{
\begin{tabular}{@{}l@{}l@{\hskip 2em}rr@{\hskip 3em}rr@{}}
\toprule
& \multirow{2}{*}{\textbf{Dataset}} & \multicolumn{2}{c@{\hskip 3em}}{\textbf{Original}} & \multicolumn{2}{c}{\textbf{Long-form}} \\
\cmidrule(r{3em}){3-4} \cmidrule{5-6}
 & & \multicolumn{1}{c}{\textbf{Size}} & \multicolumn{1}{c@{\hskip 3em}}{\textbf{Length}} & \multicolumn{1}{c}{\textbf{Size}} & \multicolumn{1}{c}{\textbf{Length}} \\
\midrule
\parbox[t]{2.5mm}{\multirow{4}{*}{\rotatebox[origin=c]{90}{{\footnotesize GigaSpeech}}}}\ldelim\{{4}{0.4cm}
& Train (M) & 999:56 & 4.0 & 1077:25 & 11.8 \\
& Train (200h) & 195:26 & 4.0 & 223:10 & 295.0 \\
& Dev & 11:50 & 6.6 & 10:29 & 1510.9 \\
& Test & 39:39 & 5.7 & 39:18 & 1088.2 \\
\midrule
\parbox[t]{2.5mm}{\multirow{5}{*}{\rotatebox[origin=c]{90}{{\footnotesize TED-LIUM}}}}\ldelim\{{5}{0.4cm}
& Train & 453:48 & 6.1 & \makecell[r]{441:59\\(514:23)} & \makecell[r]{64.0\\(827.4)} \\
& Dev & 1:35 & 11.3 & \makecell[r]{1:31\\(1:35)} & \makecell[r]{459.3\\(815.3)} \\
& Test & 2.37 & 8.2 & \makecell[r]{2:24\\(2:42)} & \makecell[r]{576.1\\(972.7)} \\
\midrule
\parbox[t]{2.5mm}{\multirow{3}{*}{\rotatebox[origin=c]{90}{{\footnotesize VoxPopuli}}}}\ldelim\{{3}{0.4cm}
& Train & 536:08 & 10.6 & 1111:46 & 143.7  \\
& Dev & 5:06 & 10.5 & 7:31 & 129.5\\
& Test & 5:04 & 9.9 & 18:01 & 108.5 \\
\bottomrule
\end{tabular}}
\end{table}

Table~\ref{tab:stats} provides summary statistics contrasting the original and reconstituted long-form versions of the corpora.
Since GigaSpeech reconstitution is simply linking, the extra corpus size is entirely between-segment silence. 
We created two long-form versions: M and 200h.
The former is obtained by linking segments of the original GigaSpeech-M (GS-M), which is approximately 1000 hours.
Since GS-M is a \textit{random} subset of GS-XL, it does not have many consecutive segments; as a result, the reconstituted long-form version only has an average length of 11.8s. 
To solve this issue, we created GS-200h specifically out of the longest consecutive segments in GS-XL, resulting in segments that are at least 240s. 
The \texttt{dev} and \texttt{test} have no missing references; therefore, their long-form versions are the full recordings.

For TED-LIUM, we show statistics for both the \textit{linked} and \textit{expanded} versions of the reconstituted data (we used the former for ASR experiments in Section~\ref{sec:asr}). 
The increase in partition size for the expanded corpora results primarily from inclusion of inter-segment silence, not new references.

For VoxPopuli, expansion resulted in the addition of a substantial amount of new data.
%
Compared to GS and TL, long-form segments are relatively short because we used the original paragraph segmentation.

\vspace{-1em}
\subsection{Alignment}
\vspace{-0.5em}







In addition to extended transcriptions, we also provide word-level timestamps obtained through forced alignments.
For the linked corpora, we used a HMM-GMM model trained using Kaldi~\cite{Povey2011TheKS} to align the original short segments.
For the expanded corpora, we modified the Fairseq aligner~\cite{pratap2023scaling} to provide start and end times and accompanying scores for each word. 
This aligner was run on the complete TED-LIUM talks and the VoxPopuli paragraphs. 
We used these word-level timestamps to create fixed-length chunks for training (e.g., 15 or 30 seconds) by concatenating subsequent words until the segment exceeded the specific length.
We have supplied these alignments in our distribution to enable users to experiment with other segment lengths or dynamic segmentation.

\vspace{-1em}
\section{ASR benchmarks}
\label{sec:asr}
\vspace{-0.5em}

In conventional ASR, audio features for a segmented utterance $\mathbf{X} \in \mathbb{R}^{T\times F}$ are provided as input to the system, and we are required to predict the transcript $\mathbf{y} = (y_1,\ldots,y_U)$, where $y_u \in \mathcal{Y}$ denotes output units such as graphemes or word-pieces. 
%
ASR systems search for $\hat{\mathbf{y}} = \text{arg}\max_{\mathbf{y}}P(\mathbf{y}|\mathbf{X})$, often in a constrained search space using greedy or beam search.


\vspace{-1em}
\subsection{Models}
\vspace{-0.5em}


\textbf{Neural transducers} are trained by minimizing the conditional log-likelihood, by marginalizing over the set of all alignments $\mathbf{a} \in \bar{\mathcal{Y}}^{T+U}$, where $\bar{\mathcal{Y}} = \mathcal{Y}\cup \{\phi\}$ and $\phi$ is called the blank label~\cite{Graves2012SequenceTW}. Formally, $P(\mathbf{y}|\mathbf{X}) = \sum_{\mathbf{a}\in \mathcal{B}_{\mathrm{RNNT}}^{-1}(\mathbf{y})}P(\mathbf{a}|\mathbf{X})$,
%
%
where $\mathcal{B}_{\mathrm{RNNT}}$ removes the blank token.
%
The probability $P(\mathbf{x}|\mathbf{X})$ is computed by factoring the parameters into an encoder, a prediction network, and a joiner.
Since transducers are trained using alignment between $\mathbf{X}$ and $\mathbf{y}$, they do not need to model end of sequence explicitly.
This \textit{frame-synchronous} behavior may result in more robustness to train-test length mismatch.
Furthermore, it also allows the estimation of token-level time-stamps at inference.

Conversely, we also trained a \textit{label-synchronous} \textbf{attention-based encoder-decoder} (AED) in the joint CTC-attention framework~\cite{Kim2016JointCB}. 
The attention head is trained with a label-wise cross-entropy loss, whereas the CTC head is trained with a sequence-level alignment-free criterion~\cite{Graves2006ConnectionistTC}.
%

\vspace{-1em}
\subsection{Overlapped chunk-wise decoding}
\vspace{-0.5em}

We follow the overlapped chunk-wise decoding strategy~\cite{chiu2019comparison}. 
Given a long recording, we chunk it into fixed-length segments of size $\ell_{\mathrm{ch}}$, and extend them by an additional $\ell_{\mathrm{ex}}$ on each side to avoid edge effects.
%
%
These segments are decoded using the transducer/AED model to obtain time-stamped tokens. We discard the edge tokens which belong to the extra regions in each segment. 
Finally, we concatenate the segment-level tokens to obtain the transcript for the recording. 
Unlike \cite{Kang2021PartiallyOI}, we do not need to align the overlapped regions of consecutive segments, but the models are required to estimate token-level time-stamps.

\vspace{-1em}
\subsection{Long-form inference with attention decoding}
\vspace{-0.5em}

We make two changes to the standard attention decoding paradigm to enable long-form inference. First, to combat this problem of high deletion rates on longer segments, we remove short hypotheses after beam search - any hypotheses with 10 or more tokens fewer than the longest hypothesis is removed from consideration. We use this setting for all AED decoding results. Second, because token-level time-stamps are required for the overlapped inference method described above, we obtain these from a constrained forward pass with the CTC head after decoding with the attention head.

\vspace{-1em}
\subsection{Experimental Setup}
\vspace{-0.5em}

For our experiments with neural transducers, we modified the standard Zipformer-transducer recipe in icefall\footnote{\url{https://github.com/k2-fsa/icefall}}, trained using the pruned RNN-T loss~\cite{Kuang2022PrunedRF}. The encoder consists of 6 Zipformer blocks~\cite{zipformer}, which are subsampled by up to 8x, and contain multiple self-attention layers (with shared attention weights). The prediction network is a 1D-convolutional layer with bigram context. We used greedy search for the chunk-wise decoding strategy. AEDs are implemented as part of a joint CTC/attention model using the Wenet toolkit\footnote{\url{https://github.com/wenet-e2e/wenet}}. They consist of a 12-layer conformer~\cite{gulati2020conformer} encoder and 6-layer bidirectional transformer decoder, although only the 3 forward decoder layers are used for inference. We use beam search for all decoding conditions. For both models, 80-dim log Mel filter-banks are used as acoustic input, and the output units are BPEs. We trained the models using SpecAugment~\cite{Park2019SpecAugmentAS} on the original segments as well as on a combined training set containing the original and the long-form segments. Due to GPU memory constraints, the long-form partition was split into 30s segments. We evaluated the models on the original segments to compare word error rate (WER) performance with oracle segmentation, and then on the reconstituted long-form sets to measure robustness to train-test mismatch.



\vspace{-1em}
\subsection{Results}
\vspace{-0.5em}

\begin{table}[t]
\centering
\caption{WER results on original test sets.}
\label{tab:orig_asr}
\adjustbox{max width=\linewidth}{
\begin{tabular}{@{}lrrrrrrr@{}}
\toprule
\multirow{2}{*}{\textbf{Model}} & \multirow{2}{*}{\makecell[r]{\textbf{Size}\\\textbf{(M)}}} & \multicolumn{2}{c}{\textbf{TL}} & \multicolumn{2}{c}{\textbf{GS}} & \multicolumn{2}{c}{\textbf{VP}} \\
\cmidrule(r{5pt}){3-4} \cmidrule(l{4pt}){5-6} \cmidrule(l{4pt}){7-8}
 &  & \textbf{Dev} & \textbf{Test} & \textbf{Dev} & \textbf{Test} & \textbf{Dev} & \textbf{Test} \\ 
\midrule
Transducer & 65.5 & 6.38 & 5.86 & 14.49 & 13.98 & 8.03 & 8.29 \\
AED & 109.8 & 9.11 & 8.48 & 15.34 & 15.31 & 13.63 & 14.07 \\
\bottomrule
\end{tabular}}
\end{table}


\begin{table}[t]
\centering
\caption{WER results on long-form evaluation. $^\dagger$Long-form training for VoxPopuli contains non-verbatim transcripts.}
\label{tab:lf_asr}
\adjustbox{max width=\linewidth}{
\begin{tabular}{@{}llrrrrrr@{}}
\toprule
\multirow{2}{*}{\textbf{Model}} & \multirow{2}{*}{\begin{tabular}{@{}c}\textbf{Training}\\\textbf{data}\end{tabular}} & \multicolumn{2}{c}{\textbf{TL}} & \multicolumn{2}{c}{\textbf{GS}} & \multicolumn{2}{c}{\textbf{VP}$^{\dagger}$} \\
\cmidrule(r{5pt}){3-4} \cmidrule(l{4pt}){5-6} \cmidrule(l{4pt}){7-8}
 & & \textbf{Dev} & \textbf{Test} & \textbf{Dev} & \textbf{Test} & \textbf{Dev} & \textbf{Test} \\ 
\midrule
\multirow{2}{*}{Transducer} & Original & 7.02 & 6.08 & 17.09 & 17.06 & 16.37 & 19.33 \\
 & + Long-form & 6.25 & 5.71 & 16.21 & 16.36 & 26.18 & 29.84 \\
\midrule
\multirow{2}{*}{AED} & Original & 60.58 & 62.77 & 45.05 & 45.50 & 34.21 & 39.06 \\
 & + Long-form & 18.88 & 23.89 & 20.17 & 20.78 & 27.83 & 33.02\\
\bottomrule
\end{tabular}}
\end{table}

\begin{figure}[t]
\begin{subfigure}{0.32\linewidth}
\centering
\includegraphics[width=1.05\linewidth]{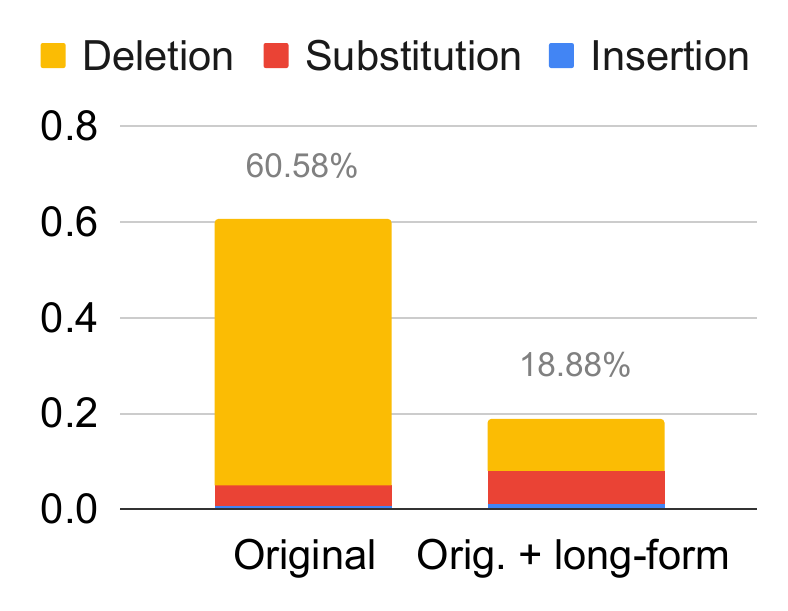}
\vspace{-1.5em}
\caption{TED-LIUM}
\label{fig:tl}
\end{subfigure}
\begin{subfigure}{0.32\linewidth}
\centering
\includegraphics[width=1.05\linewidth]{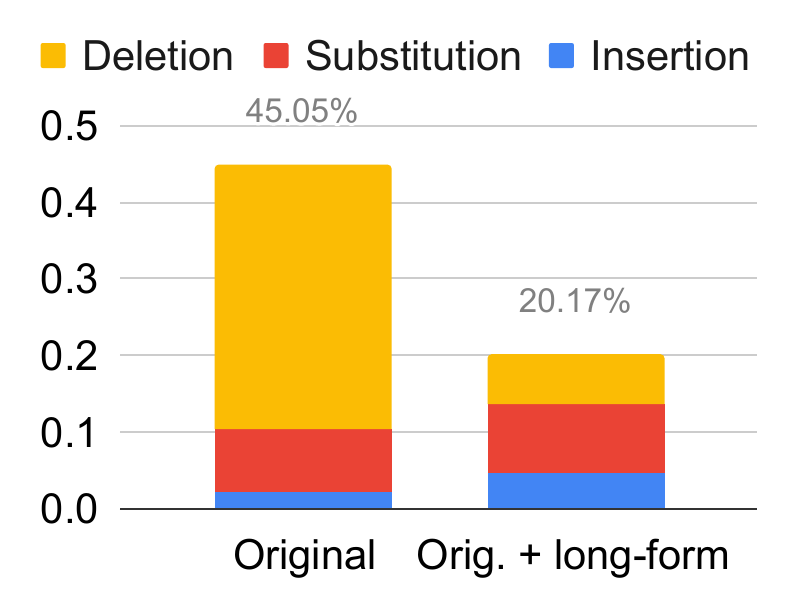}
\vspace{-1.5em}
\caption{GigaSpeech}
\label{fig:gs}
\end{subfigure}
\begin{subfigure}{0.32\linewidth}
\centering
\includegraphics[width=1.05\linewidth]{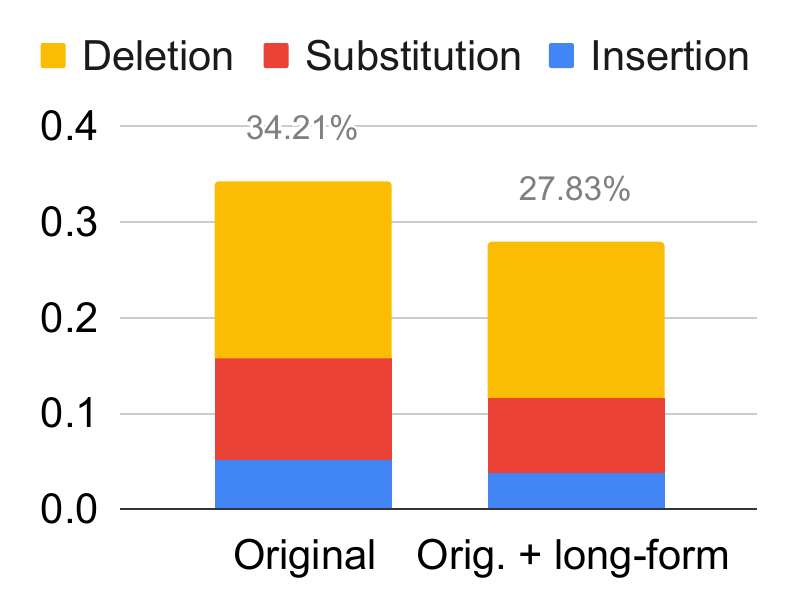}
\vspace{-1.5em}
\caption{VoxPopuli}
\label{fig:vp}
\end{subfigure}\hfill
\vspace{-0.5em}
\caption{Error rates on the \texttt{dev} sets for AED long-form inference.}
\label{fig:aed_lf_error_rates}
\vspace{-0.5em}
\end{figure}

Table~\ref{tab:orig_asr} shows the performance of the different model types when trained and evaluated on the original segments. 
Across all three corpora, the transducer model performed best, but both model types gave reasonable results. 
However, when these models were used to decode the reconstituted long-form data, their performance varied significantly, as seen in the ``Original'' rows of Table~\ref{tab:lf_asr}.
%
As expected, the transducer model degraded only slightly (e.g., 6.38\%$\rightarrow$7.02\% on TED-LIUM dev), whereas \textbf{AED degraded significantly} on all test suites, driven predominantly by high deletion rates. 
The breakdown of AED WERs into insertion, substitution, and deletion errors can be seen in Figure~\ref{fig:aed_lf_error_rates}.

%

Table~\ref{tab:lf_asr} also shows the results for long-form training using the updated \texttt{train} sets. 
For both TED-LIUM and GigaSpeech, this training led to small improvements in transducer performance and large improvements in AED performance, although the transducer was still significantly better than the AED for long-form inference. 
From Fig.~\ref{fig:aed_lf_error_rates}, we see that long-form training resulted in \textbf{large reductions in the deletion rate}, leading to better performance.

The inclusion of the long-form VoxPopuli data degraded performance for the transducer model, and only improved the AED model slightly. 
This is likely due to our being overly permissive with the included data. 
Some of the VoxPopuli references in the original transcripts were heavily edited (i.e., non-verbatim transcription), including reordering of words. 
Since the transducer model assumes monotonic alignments, training with such transcripts could potentially deteriorate the model. 
Future work is needed to find the appropriate balance between including additional data needed for long-form training and rejecting low-quality references. 
Alternatively, recently proposed techniques such as bypass temporal classification~\cite{gao2023bypass}, which allow training with imperfect transcripts, could be explored for making the best use of this data.

\vspace{-1em}
\section{Conclusion}
\label{sec:conclusion}
\vspace{-0.5em}

%
In this work, we released updated long-form versions of three popular English datasets --- TED-LIUM, GigaSpeech, and VoxPopuli.
This was achieved using a general ``reconstitution'' recipe comprising linking and expansion stages.
To accompany this release, we presented baseline results using two commonly used models, transducers and AEDs. 
Across all three datasets, we demonstrated that transducers are more robust than AEDs to the train/test mismatch, when trained on segmented utterances. 
Finally, we showed that a simple strategy of combining original and long-form segments for training is effective at reducing the performance gap. 
Nevertheless, more research into training and modeling strategies is required to make long-form ASR robust in real scenarios, and we believe our public benchmarks would be important to measure progress.

\vspace{+1em}
{\small
\begin{spacing}{0.85}
\noindent
\textbf{Acknowledgments.} This work was started during JSALT 2023, hosted at Le Mans University, France, and sponsored by Johns Hopkins University with unrestricted gifts from Amazon, Facebook, Google, and Microsoft. D.R. acknowledges funding by NSF CCRI Grant No. 2120435 and a JHU-Amazon AI2AI fellowship.
\end{spacing}
}
\clearpage
\small
\setstretch{0.9}
\bibliographystyle{IEEEbib-abbrev}
\bibliography{refs}

\end{document}